\documentclass[preprint,1p]{elsarticle}

\usepackage{amssymb}
\usepackage{color}
\usepackage{xspace}
\usepackage{pdfwidgets}
\usepackage{enumerate}
\usepackage{color}
\usepackage{bm}
\usepackage{graphicx}
\usepackage{here}
\usepackage{amsmath}

\newcommand{\argmin}{\mathop{\rm arg~min}\limits}

\journal{Neurocomputing}

\begin{document}

\begin{frontmatter}



\title{R2-Diff: Denoising by diffusion as a refinement of retrieved motion for image-based motion prediction}


\author{Takeru Oba}
\ead{sd21502@toyota-ti.ac.jp}
\author{Norimichi Ukita}
\ead{ukita@toyota-ti.ac.jp}

\affiliation{organization={Toyota Technological Institute},
            country={Japan}}



\begin{abstract}
Image-based motion prediction is one of the essential techniques for robot manipulation. Among the various prediction models, we focus on diffusion models because they have achieved state-of-the-art performance in various applications. In image-based motion prediction, diffusion models stochastically predict contextually appropriate motion by gradually denoising random Gaussian noise based on the image context. While diffusion models are able to predict various motions by changing the random noise, they sometimes fail to predict a contextually appropriate motion based on the image because the random noise is sampled independently of the image context. To solve this problem, we propose R2-Diff. In R2-Diff, a motion retrieved from a dataset based on image similarity is fed into a diffusion model instead of random noise. Then, the retrieved motion is refined through the denoising process of the diffusion model. Since the retrieved motion is almost appropriate to the context, it becomes easier to predict contextually appropriate motion. However, traditional diffusion models are not optimized to refine the retrieved motion. Therefore, we propose the method of tuning the hyperparameters based on the distance of the nearest neighbor motion among the dataset to optimize the diffusion model for refinement. Furthermore, we propose an image-based retrieval method to retrieve the nearest neighbor motion in inference. Our proposed retrieval efficiently computes the similarity based on the image features along the motion trajectory. We demonstrate that R2-Diff accurately predicts appropriate motions and achieves high task success rates compared to recent state-of-the-art models in robot manipulation.
\end{abstract}

\end{frontmatter}


\section{Introduction}
\label{section:intro}

\begin{figure}[h]
  \begin{center}
    \includegraphics[width=1.0\columnwidth]{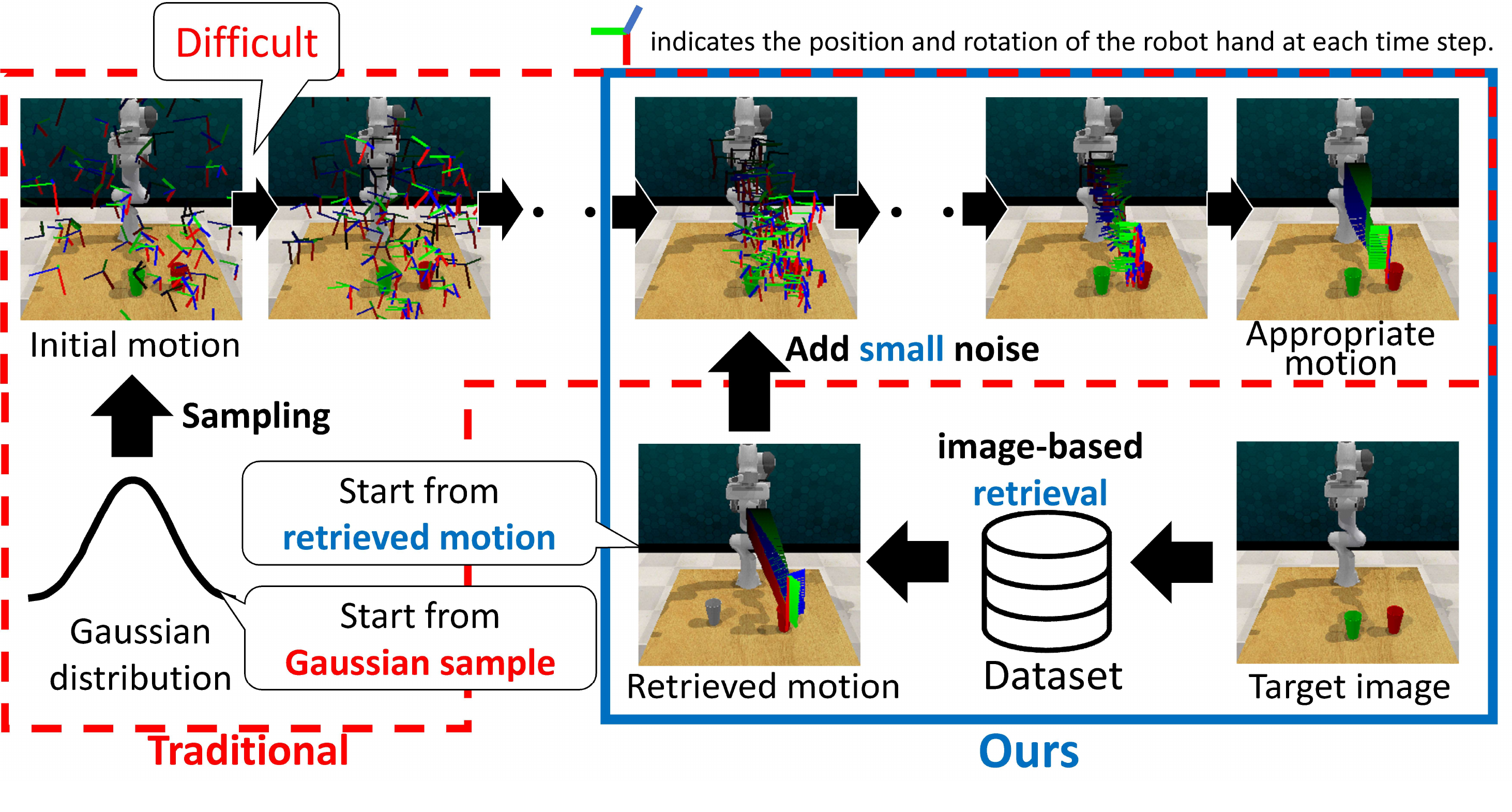}
    \caption{Overview of the proposed diffusion process. The traditional diffusion process starts denoising from Gaussian noise. In contrast, our diffusion starts denoising from noised retrieved motion and avoids difficult denoising in the early steps.}
    \label{fig:overview of proposal}
  \end{center}
\end{figure}

Motion prediction~\cite{bojarski2016end,codevilla2019exploring,zhang2018deep,pari2021surprising, oba2023data}, also referred to as motion planning or imitation learning, is one of the essential techniques for robot manipulation. Since motion prediction requires understanding the relationships between an environment and a motion, image-based motion prediction approaches utilize images as a representation of the environment and learn complex relationships by neural networks~\cite{zhang2018deep, pari2021surprising, oba2023data}. 
One of the difficulties in image-based motion prediction is the stochasticity of motions, i.e., various motions could accomplish a task in a certain environment. This stochasticity makes it difficult to predict the correct motion with a deterministic model because the deterministic model often predicts an averaged motion over all possible motions appropriate to the image context. Therefore, stochastic models are often used for motion prediction~\cite{oba2023data, chi2023diffusion, zhang2020f}.

Recently, diffusion models~\cite{ho2020denoising, song2020score, nichol2021improved}, one of the stochastic models, have achieved the state-of-the-art performance in various domain benchmarks (e.g., image synthesis~\cite{meng2021sdedit, kawar2022denoising}, motion synthesis~\cite{tevet2022human, zhang2023remodiffuse}, motion prediction~\cite{chi2023diffusion}, etc.).
In image-based motion prediction, the diffusion model predicts a motion conditioned on image context by gradually denoise an initial motion, as shown in the area enclosed by the red dotted lines in Fig.~\ref{fig:overview of proposal}.
This gradual denoise approach stabilizes the learning of stochasticity and improves prediction accuracy.
However, this denoise approach has two problems.
The first problem is the randomness of the initial motion.
Basically, in traditional diffusions, the initial motion is \textit{randomly} sampled from a Gaussian distribution, specifically a standard normal distribution.
While this randomness allows the diffusion model to sample various motions, large randomness sometimes leads the model to fail to predict the motion appropriate to the image context.
The second problem is the difficulty of denoising in early denoising steps.
As we mentioned before, the diffusion model denoises a motion randomly sampled from a Gaussian distribution so that it gradually becomes appropriate to the image context. However, the early stages of denoising are difficult because the motion is quite different from the appropriate motion, as shown on the left side of Fig.~\ref{fig:overview of proposal}. This difficulty deteriorates both training and inference of motion prediction.

To solve these problems, we propose R2-Diff. In R2-Diff, the aforementioned motion prediction from a random sample (i.e., Gaussian sampling) is replaced by motion refinement from a feasible initial motion, as shown in the blue part of Fig.~\ref{fig:overview of proposal}. This feasible motion is retrieved from the training dataset based on the similarity to the test image. While this retrieval reduces the variety of motions because the randomness of the initial motion is reduced, it improves the accuracy of the prediction because the retrieved motion is more feasible than a random motion.


Furthermore, we also propose a method for tuning the noise scale for the R2-Diff. The noise scale is an important factor for prediction accuracy in diffusion models. Traditionally, since the initial motion is randomly sampled from the Gaussian distribution, the noise scale has been manually tuned to fit the Gaussian distribution at the last noise step~\cite{ho2020denoising, chi2023diffusion}. In contrast, the initial motion is retrieved from the dataset in R2-Diff. Since the retrieved motion is closer to the ground-truth motion, the traditional noise scale at the early denoising step (shown on the left side of Fig.~\ref{fig:overview of proposal}) is too large and unnecessary for R2-Diff. Training with this unnecessarily-large noise could prevent the model from learning the motion accurately. To avoid this training, our tuning method computes the distance of the nearest neighbor motion among the training dataset, assuming that the nearest neighbor motion is retrieved in inference. Then, the noise scale is adjusted to fit this distance. Our tuning allows us to find the appropriate noise scale for any task without training or manual tuning.

Finally, we propose an image-based motion retrieval method for accurately retrieving a motion close to the ground-truth motion. To retrieve the motion, the similarity to the test image is computed through the image features. Then, the motion corresponding to the most similar image in the training dataset is retrieved. Traditional retrieval methods compute the distance between image features downscaled uniformly by convolution or pooling. Since such uniform downscaling is independent of the motion, the image features often include useless and interrupting information (e.g., background) that disturbs our retrieval task. To address this problem, we employ the Spatially-aligned Temporal Embedding features (STE features) proposed in~\cite{oba2023data}. STE extracts image features along a 2D motion trajectory on the image plane. This motion-dependent feature extraction removes the disturbing information if appropriate motions are given. However, while the appropriate motion for images in the training dataset is known, that of the test image is unknown. Therefore, we propose to extract the test image feature based on each ground-truth motion in the training dataset. When a motion in the training data is almost appropriate to a test image, image features of the target and training image obtained along that motion trajectory should be almost the same. Therefore, the appropriate motion can be retrieved by extracting image features along the motion trajectory corresponding to each training image and calculating the error between the obtained image features.


The novel contributions proposed in this paper are as follows:
\begin{itemize}
\item Our retrieval and refinement diffusion model (R2-Diff) accurately predicts the motion through denoising as refinement from a retrieved motion instead of denoising from a motion randomly sampled from a Gaussian distribution.
\item The noise scale of R2-Diff is automatically computed based on the nearest neighbor search among the training dataset to avoid difficult training harms the model.
\item The proposed retrieval allows us to use STE to accurately retrieve the appropriate motion from the image similarity by extracting important features located in different regions for each image.
\end{itemize}

\section{Related Work}
\subsection{Motion prediction}
Motion prediction~\cite{bojarski2016end,codevilla2019exploring,zhang2018deep,pari2021surprising, oba2023data} allows us to achieve a variety of real-world systems, such as robotic process automation~\cite{zhang2018deep,pari2021surprising, oba2023data}.
The motion is predicted so that the robot accomplishes a task from the observed data (e.g., image).
The difficulty of motion prediction is as follows: (1) time-series data processing, (2) stochasticity of motion, and (3) conditioning on observation.

The first difficulty is that motion is time-series data. As well as other time-series data forecasting (e.g., language~\cite{lewis2020bart, leiter2023chatgpt}, audio signal~\cite{ruan2022mm, leng2022binauralgrad}, stock price~\cite{VIJH2020599}), RNNs~\cite{hochreiter1997long, cho2014learning} and Transformers~\cite{aksan2021spatio, giuliari2021transformer, kim2021transformer} are often used for motion prediction because these models efficiently encode temporal information.

The second problem is caused in a deterministic model because various motions that can accomplish the task tend to be averaged, as explained in Section~\ref{section:intro}. The averaged motion is not guaranteed to accomplish the task. To deal with this stochasticity, stochastic models such as GANs~\cite{barsoum2018hp, ho2016generative, hernandez2019human}, VAEs~\cite{aliakbarian2021contextually, wang2017robust, cao2020long}, EBMs~\cite{oba2023data}, and Diffusions~\cite{chi2023diffusion} are utilized to learn multiple possible motions. Recently, diffusion models have achieved state-of-the-art performance in various benchmarks among these models because the gradual denoising applied in diffusion models is more stable than other stochastic models. However, while these stochastic modes are effective for predicting the various motions, this randomness sometimes leads to inaccurate predictions, even in diffusion models.

The third problem is that the motion prediction should be properly conditioned on the observation to predict the motion appropriate to the observation context. While concatenating observation features with motion features is the simplest way for conditioning~\cite{zhang2018deep, finn2017one}, transformers have recently been used for conditioning various observation data~\cite{chen2021decision, tevet2022human}. Transformers allow us to efficiently condition observation features through attention mechanisms weighting the important features for conditioning. Furthermore, unlike concatenation, transformers allow the length of the observation sequence to be changed during inference. However, transformers have a computational cost problem. For example, if an image is fed into the model without downscaling, the cost of the transformer becomes huge. Spatially aligned Temporal Encoding (STE)~\cite{oba2023data} is one of the ways to solve this problem by extracting image features in important regions computed by projecting a motion onto the image plane. Therefore, we employ a transformer and STE for the conditioning of images.

\subsection{Diffusion}
\label{section:related diffusion}

\begin{figure}[h]
  \begin{center}
    \includegraphics[width=1.0\columnwidth]{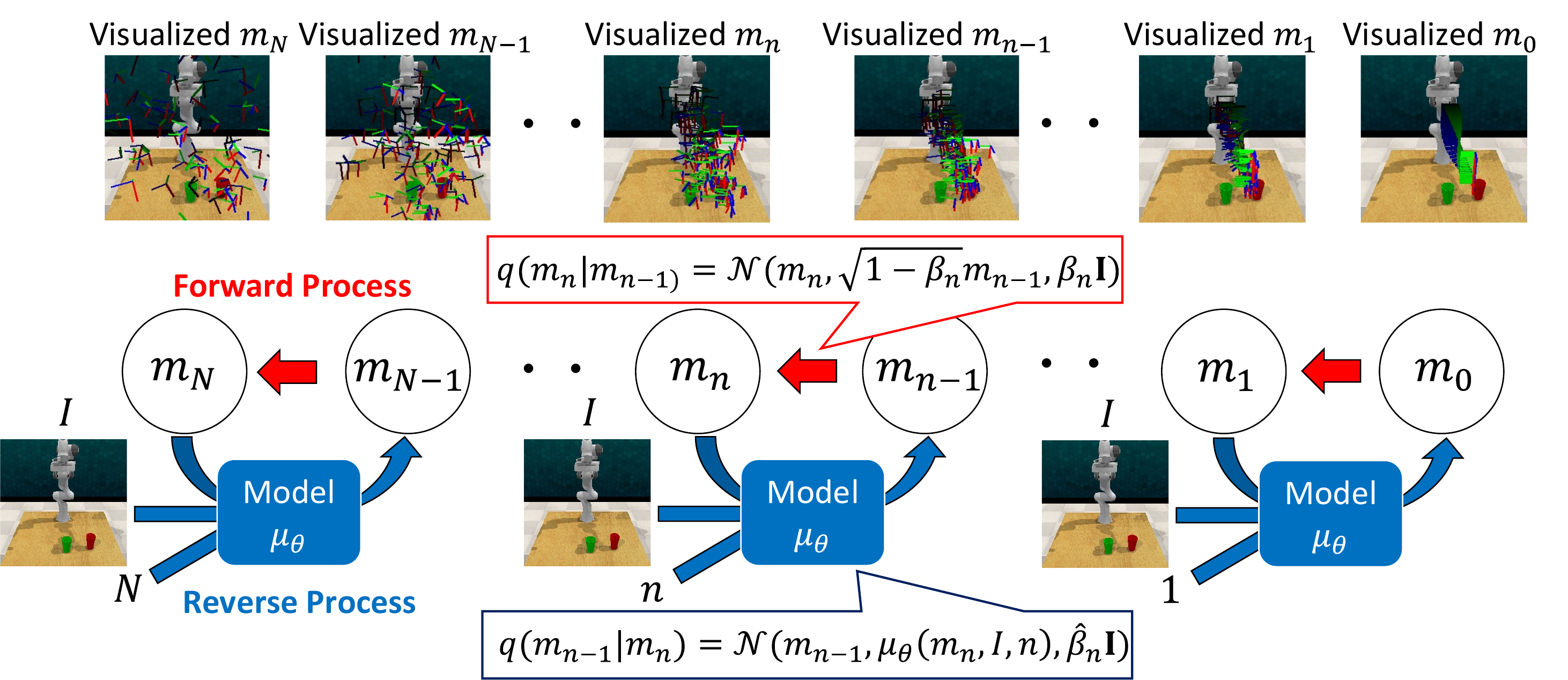}
    \caption{Overview of the traditional diffusion process. The upper part of this figure visualizes the noised motion superimposed on the image. The red, blue, and green lines represent the motion of the robot hand. The intersections and directions of these lines are the position and rotation of the robot hand at each motion timestep, respectively.}
    \label{fig:traditional diffusion}
  \end{center}
\end{figure}

Diffusion models have been applied to various tasks, such as image synthesis~\cite{meng2021sdedit, kawar2022denoising}, motion synthesis~\cite{tevet2022human, zhang2023remodiffuse}, audio signal~\cite{ruan2022mm, leng2022binauralgrad}, 3D modeling~\cite{poole2022dreamfusion, wang2022score}, shared autonomy~\cite{yoneda2023diffusha}, etc. We briefly introduce the formulation of diffusion models based on our motion prediction setting. Figure~\ref{fig:traditional diffusion} shows the overview of the diffusion process. Diffusion models have forward and reverse processes. The forward process shown in the red arrows in Fig.~\ref{fig:traditional diffusion} can be considered a noising process; the reverse process shown in the blue arrows in Fig.~\ref{fig:traditional diffusion} is a denoising process. We will introduce the forward process first.
Suppose that $\bm{m}_0$ is a target of prediction. In our case, $\bm{m}_0$ is a motion appropriate to an image $I$. By adding Gaussian noise, $\bm{m}_0$ gradually becomes noisy. This noisy motion at noise step $n \in \{1,\cdots,N\}$ is denoted by $\bm{m}_n$. $N$ is the maximum number of noise steps. This noising process is formulated as follows with $\beta_n \in (0, 1)$, which is the noise variance at noise step $n$,

\begin{equation}
    q(\bm{m}_n \mid \bm{m}_{n-1}) := \mathcal{N}(\bm{m}_n; \sqrt{1 - \beta_n} \bm{m}_{n-1}, \beta_n \bf{I}),
\end{equation}
$\bf{I}$ denotes identity matrix. $q(\bm{m}_n \mid \bm{m}_{n-1})$ denotes the transition probability from $\bm{m}_{n-1}$ to $\bm{m}_{n}$. This transition is implemented as follows:
\begin{equation}
    \bm{m}_n = \sqrt{1 - \beta_{n}} \bm{m}_{n-1} + \sqrt{\beta_{n}} \epsilon, 
\end{equation}
where $\epsilon \sim \mathcal{N}(0, \bf{I})$.
Since the noise at each noise step is independent, the noised motion $\bm{m}_n$ at the arbitrary noise step $n$ is computed from $\bm{m}_0$ as follows:

\begin{equation}
    q(\bm{m}_n \mid \bm{m}_{0}) = \mathcal{N}(\bm{m}_n; \sqrt{\bar{\alpha}_{n}} \bm{m}_{0}, (1 - \bar{\alpha}_{n}) \bf{I}), 
\end{equation}
where $\bar{\alpha}_n := \prod_{i=1}^{n} (1 - \beta_i)$.
The distribution of $\bm{m}_n$ becomes the standard normal distribution when $\bar{\alpha}_n$ is $1$. Since sampling from the standard normal distribution is easy, $\beta_1,\cdots,\beta_{N}$ are determined so that $\bar{\alpha}_N$ close to $1$ at noise step $N$.
In inference, $\bm{m}_N$ is randomly sampled from the standard normal distribution and gradually denoised from $N$ to $0$ to predict the motion by iterating the reverse process.

While the forward process is formulated based on the Gaussian distribution, the reverse process is also formulated based on the Gaussian distribution as follows,

\begin{equation}
    q(\bm{m}_{n-1} \mid \bm{m}_{n}) = \mathcal{N}(\bm{m}_{n-1}; \mu_{\theta}(\bm{m}_n, I, n), \hat{\beta_{n}} \bf{I}),
    \label{equation: reverse process}
\end{equation}

\begin{equation}
     \mu_\theta\left(\bm{m}_n, I, t\right)=\frac{1}{\sqrt{\alpha_n}}\left(\bm{m}_n-\frac{\beta_n}{\sqrt{1-\bar{\alpha}_n}} \epsilon_\theta\left(\bm{m}_n, I, t\right)\right), \label{equation: mu}
\end{equation}

\begin{equation}
    \hat{\beta_{n}} := \frac{1 - \bar{\alpha}_{n-1}}{1 - \bar{\alpha}_{n}}, \label{equation: beta}
\end{equation}
where $\epsilon_{\theta}(\bm{m}_n, I, n)$ is predicted noise from a neural network $\epsilon_{\theta}$.
See~\cite{ho2020denoising} for a derivation to Eqs. (\ref{equation: reverse process}) $\sim$ (\ref{equation: beta}).
Since our goal is predicting the motion that accomplishes the task in a given environment represented by an image $I$, $I$ is additionally fed into the model in our case. The neural network $\epsilon_{\theta}$ is trained to predict the noise $\epsilon$ from the noisy motion $\bm{m}_n$, the image $I$, and denoising step $n$ by minimizing the following loss:

\begin{equation}
    L_{\rm simple}=E_{n, \bm{m}_0, I, \epsilon}\left\|\epsilon-\epsilon_\theta\left(\bm{m}_n, I, n\right)\right\|^2. \label{equation: loss}
\end{equation}

\section{Problem setting}
\label{section:problem setting}

\begin{figure}[h]
  \begin{center}
    \includegraphics[width=1.0\columnwidth]{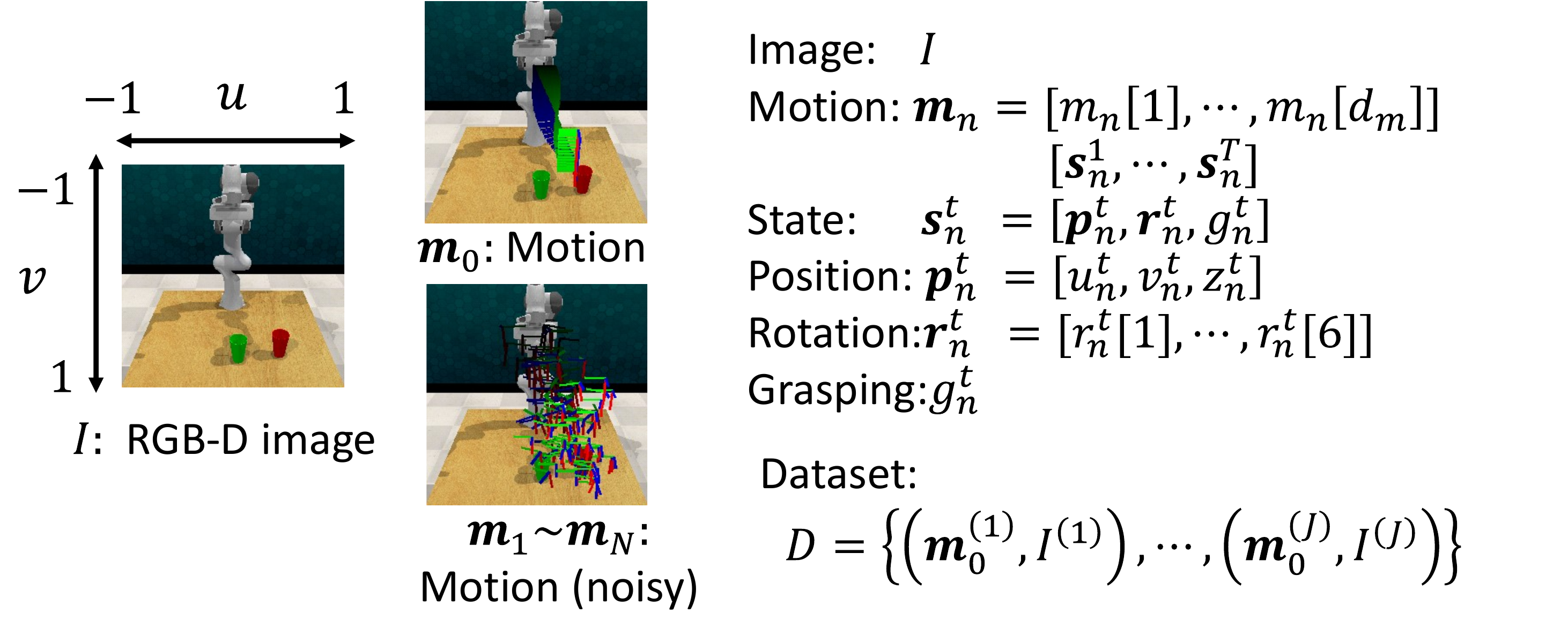}
    \caption{The list of variables explained in Section~\ref{section:problem setting}.}
    \label{fig:setting}
  \end{center}
\end{figure}

Our goal is to predict the motion $\bm{m}_0$ from a given image $I$ representing the environment. $\bm{m}_0$ is predicted by gradually denoising the noised motion $\bm{m}_n$. Figure~\ref{fig:setting} shows the details of $I$ and $\bm{m}_n$ in our problem. $I$ is an RGB-D image in which the robot and its surrounding environment are observed, as shown in Fig.~\ref{fig:setting}. $\bm{m}_n$ is a $d_m$-dimensional vector representing the motion of the robot at noise step $n$. The $d$-th element of an arbitrary vector $\bm{x}$ is denoted by $\bm{x}[d]$ (e.g., $\bm{m}_n[d]$). $\bm{m}_n$ is comprised of the states of a robotic hand from time $1$ to $T$, where the state of the robotic hand $\bm{s}^t_n$ at time $t$ consists of its position $\bm{p}^t_n$ (3 dimensions), its rotation $\bm{r}^t_n$ (6 dimensions), and its grasping state $g^t_n$ (1 dimension). Note that the timestep ``$t$'' represents the motion timestep, which differs from the noise step ``$n$'' introduced in Section~\ref{section:related diffusion}. 
The position $\bm{p}^t_n$ is represented as a combination of the image coordinates ($u^t_n, v^t_n$) and the depth from the camera ($z^t_n$) of the robotic hand. The rotation $\bm{r}^t_n$ is represented by a six-dimensional vector, as proposed in~\cite{Zhou2019rotation}, because common representations such as Euler angles and quaternions are discontinuous and difficult to learn. The grasping state $g^t_n$ is a scalar between $0$ and $1$, representing the open (1) and closed (0) states. To actuate a robot, all joint angles of the robot at world coordinates are required. Therefore, $\bm{p}^t_0$ is converted to world coordinates with a pre-computed camera matrix, and the joint angles are computed by inverse kinematics in the robot manipulation experiment.
The training dataset denoted as $D=\{(\bm{m}^{(1)}_0, I^{(1)}),\cdots,(\bm{m}^{(j)}_0, I^{(j)}),\cdots,(\bm{m}^{(J)}_0, I^{(J)})\}$ is prepared in advance to train the model. $j \in \{1,\cdots,J\}$ and $J$ denote the index and the number of the image and motion pairs in the training dataset, respectively. The index $j$ is parenthesized to avoid confusion with timestep $t$.

\section{Model Architecture}
\label{section:model architecture}

\begin{figure}[h]
  \begin{center}
    \includegraphics[width=1.0\columnwidth]{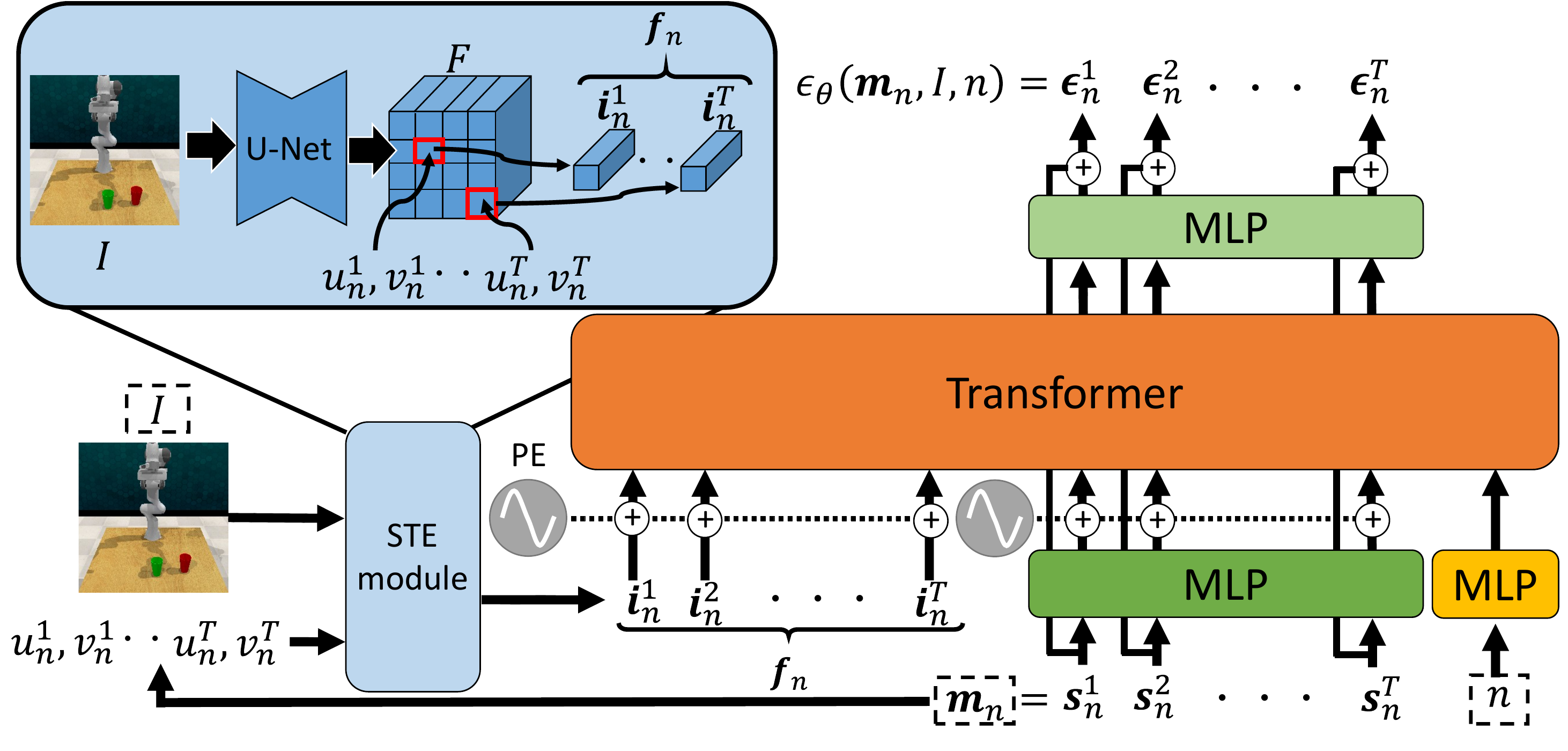}
    \caption{The architecture of our diffusion model. Our diffusion model predicts noise $\epsilon$ from noised motion $\bm{m}_{n}$, image $I$, and noise step $n$.}
    \label{fig:model}
  \end{center}
\end{figure}

Before introducing the novelty and contributions of our method, this section introduces the architecture of our diffusion model to facilitate understanding of our method. Figure~\ref{fig:model} shows the overall architecture of our diffusion model. This architecture is based on the model proposed in~\cite{tevet2022human}. While the model proposed in~\cite{tevet2022human} predicts a human motion from a sentence, our model predicts the robot motion from an image. Therefore, the image encoder is employed instead of the sentence encoder. Our image encoder, named STE module, is visualized in the bluer area in Fig.~\ref{fig:model}. STE module uses the Spatially-aligned Temporal Encoding (STE) proposed in~\cite{oba2023data} for feature extraction because it allows feature reduction without losing the precise position information. In STE, the image feature map $F$ is obtained from the U-Net-like encoder. Since $F$ is not downscaled, $F$ retains the precise position information. Then, the image feature $\bm{f}_{n} = [\bm{i}_n^{1},\cdots,\bm{i}_n^{T}]$ is extracted along the image coordinates of the robot hand ($u_{n}^{1}, v_{n}^{1},\cdots,u_n^{T},v_n^{T}$) because the scene information around the robot hand is important to judge whether the current motion $\bm{m}_n$ is appropriate for the image $I$.
The other parts of our model are almost the same as~\cite{tevet2022human}. The motion $\bm{m}_n$ and the denoise step $n$ are independently fed into the different MLP layers to obtain the features, as shown in the bottom right side of Fig.~\ref{fig:model}. Then, the image, motion, and denoise step features are fed into the transformer with sinusoidal positional encoding (PE). The motion time step $t$ is encoded by PE so that $\bm{s}_n^{t}$ and $\bm{i}_n^{t}$ at the same time-step have a high correlation. Then, features from the transformer are fed into MLP, and the noise contained in $\bm{m}_n$ is predicted as the residual of $\bm{m}_n$.

\section{Proposed Method}
\subsection{R2-Diff}
\label{section:R-diff}

\begin{figure}[h]
  \begin{center}
    \includegraphics[width=1.0\columnwidth]{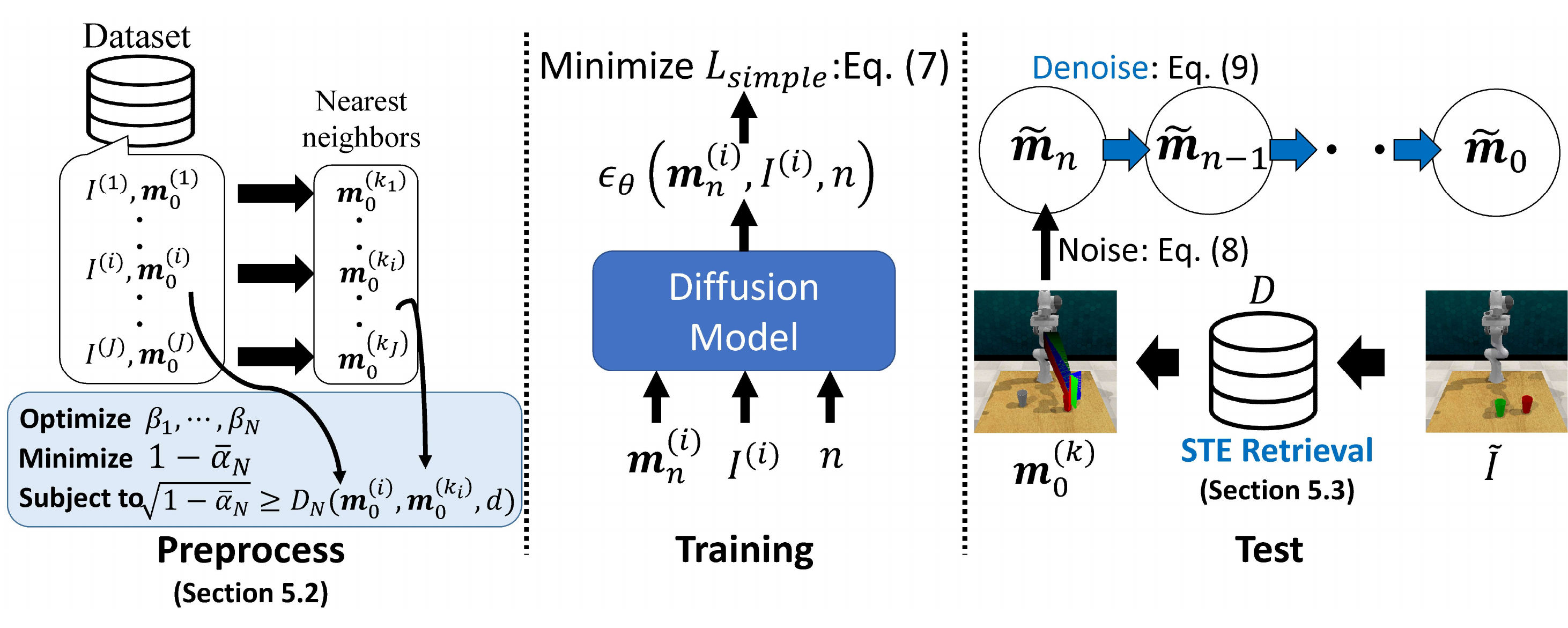}
    \caption{The overall processes of R2-Diff. The left, middle, and right sides of this figure show the preprocess, training, and test, respectively. }
    \label{fig:procedure of R2-Diff}
  \end{center}
\end{figure}

Figure~\ref{fig:procedure of R2-Diff} shows the overall process of R2-diff.
First, the noise scale hyperparameters ($\beta_{1},\cdots,\beta_{N}$) are tuned for R2-Diff, as shown on the left side of Fig.~\ref{fig:procedure of R2-Diff}. 
Since R2-Diff assume that the motion near the ground-truth motion that is almost appropriate to the image context is retrieved during inference, $\beta_{1},\cdots,\beta_{N}$ are adjusted so that the noise scale ($1 - \bar{\alpha}_{N}$) fits the distance of the nearest neighbor motion $D_N(\bm{m}_0^{(i)}, \bm{m}_0^{(k_i)}, d)$. The details of this tuning are explained in Section~\ref{section:scaling of noise}. 

The diffusion model is then trained with $\beta_{1},\cdots,\beta_{N}$ tuned for R2-Diff. The training procedure is the same as for traditional diffusion models, except for the noise scale. As shown in the middle of Fig.~\ref{fig:procedure of R2-Diff}, the noised motion $\bm{m}^{(i)}_n$, the image $I^{(i)}$, and noise step $n$ are fed into the model, and the model is optimized to predict the noise $\epsilon$ as in Eq.~(\ref{equation: loss}).

Finally, R2-Diff predicts the motion $\tilde{\bm{m}}_0$ by refining the retrieved motion $\bm{m}^{(k)}_0$ through denoising in the inference, as shown on the right side of Fig.~\ref{fig:procedure of R2-Diff}. While the nearest neighbor motion can be retrieved through the similarity of the motion in the training stage, the similarity of the motion cannot be computed because the ground truth of the motion corresponding to the test image $\tilde{I}$ is unknown. For this reason, the motion is retrieved by the similarity of the image. The details of image similarity computation are explained in Section~\ref{section:STE based retrieval}. Let $\bm{m}^{(k)}_0$ be the retrieved motion. $k$ is the index of the retrieved motion. Then, the retrieved motion $\bm{m}^{(k)}_0$ is refined to fit the test image $\tilde{I}$ context by denoising. Before the refinement, the retrieved motion is noised with Gaussian $\mathcal{N}(\bm{m}_n; \sqrt{\bar{\alpha}_{n}} \bm{m}^{(k)}_{0}, (1 - \bar{\alpha}_{n}) \bf{I})$ as follows:
\begin{equation}
    \tilde{\bm{m}}_n = \sqrt{\bar{\alpha}_{n}} \bm{m}_{0}^{(k)} + \sqrt{1 - \bar{\alpha}_{n}} \epsilon, 
\end{equation}
where $\tilde{\bm{m}}_n$ is noised retrieved motion.
This noising process addresses the domain gap between the training and inference as in~\cite{wang2022score, yoneda2023diffusha}.
Remember that the motion $\bm{m}_n$ is noised with the Gaussian in the training stage. In contrast, the retrieved motion $\bm{m}_0^{(k)}$ is not noised. Then, the model fails to refine the retrieved motion without noise due to this gap. The scale of noise is controlled by the noise step $n$. This noise step $n$ is a hyperparameter. We set $n$ to $\frac{N}{2}$ although $n$ does not significantly affect the performance, as explained in Section~\ref{section:evaluation noise step}.  The motion is refined by iterating the following equation until $n$ becomes $0$:
\begin{equation}
    \tilde{\bm{m}}_{n-1} = \mu_{\theta}(\tilde{\bm{m}}_n, \tilde{I}, n) + \sqrt{\tilde{\beta}_{n}} \epsilon,
\end{equation}
where $\mu_{\theta}(\tilde{\bm{m}}_n, \tilde{I}, n)$ and $\tilde{\beta}_{n}$ are denoted in Eqs. (\ref{equation: mu}) and (\ref{equation: beta}), respectively.

\subsection{Scaling of noise}
\label{section:scaling of noise}

This section introduces the noise scale tuning method for R2-Diff, as shown in the left side of Fig.~\ref{fig:procedure of R2-Diff}.
As explained in Section~\ref{section:related diffusion}, the distribution of noised motion $\mathcal{N}(\bm{m}_n; \sqrt{\bar{\alpha}_{n}} \bm{m}_{0}, (1 - \bar{\alpha}_{n}) \bf{I})$ is determined by $\bar{\alpha}_{n}$ computed from $\beta_{1},\cdots,\beta_{n}$. Our model is trained to denoise a noised motion $\bm{m}_n$ sampled from $\mathcal{N}(\bm{m}_n; \sqrt{\bar{\alpha}_{n}} \bm{m}_{0}, (1 - \bar{\alpha}_{n}) \bf{I})$. This means that a motion $\bm{m}_n$ outside of this distribution will not be properly denoised. Therefore, it is necessary to adjust $\bar{\alpha}_{n}$ so that retrieved motions are within this distribution. This can be accomplished by increasing the noise scale $1-\bar{\alpha}_{n}$. However, if $1-\bar{\alpha}_{n}$ is made too large, as shown on the left side of Fig.~\ref{fig:overview of proposal}, the appropriate motion $\bm{m}_{0}$ is completely obscured by the noise. As a result, denoising becomes difficult, and the model fails to predict the appropriate motion $\bm{m}_{0}$. Our proposed tuning method addresses this problem by computing the nearest-neighbor motions in the training data and minimizing $1-\bar{\alpha}_{N}$ within the range where the nearest motion is inside the distribution of the noised motion $\mathcal{N}(\bm{m}_n; \sqrt{\bar{\alpha}_{n}} \bm{m}_{0}, (1 - \bar{\alpha}_{n}) \bf{I})$ at noise step $N$.
Let $\bm{m}^{(k_i)}_0$ be the nearest neighbor motion of $\bm{m}^{(i)}_0$. The index $k_i$ is obtained by minimizing the following weighted Euclidean distance:
\begin{equation}
    k_i = \argmin_{j \in \{1, 2, \ldots, J\} \setminus \{i\}} \sqrt{\sum_{d=1}^{d_m} D_0(\bm{m}^{(i)}_0, \bm{m}^{(j)}_0, d)}, \label{equation: motion retrieval}
\end{equation}
where $D_0(\bm{m}^{(i)}_0, \bm{m}^{(j)}_0, d)$ is the weighted distance of each dimension $d$. The weighted distance $D_0(\bm{m}^{(i)}_0, \bm{m}^{(j)}_0, d)$ is represented as follows:
\begin{equation}
 D_0(\bm{m}^{(i)}_0, \bm{m}^{(j)}_0, d) = \begin{cases}
  (\bm{m}^{(i)}_0[d] - \bm{m}^{(j)}_0[d])^2, & \text{if } \bm{m}^{(i)}_0[d] \text{denotes position} \\
  w_r(\bm{m}^{(i)}_0[d] - \bm{m}^{(j)}_0[d])^2, & \text{if } \bm{m}^{(i)}_0[d] \text{denotes rotation} \\
  w_g(\bm{m}^{(i)}_0[d] - \bm{m}^{(j)}_0[d])^2  & \text{if } \bm{m}^{(i)}_0[d] \text{denotes grasping},
\end{cases}
\end{equation}
where $w_r$ and $w_g$ are weighting hyperparameters. In this paper, $w_r$ and $w_g$ are set to 0.01 and 0, respectively.
Then, let's consider the distance at noise step $N$ denoted as $D_N(\bm{m}^{(i)}_0, \bm{m}^{(k_i)}_0, d)$. Remember that the motion $\bm{m}_{0}$ is noised with the Gaussian $\mathcal{N}(\bm{m}_N; \sqrt{\bar{\alpha}_{N}} \bm{m}_{0}, (1 - \bar{\alpha}_{N}) \bf{I})$. This means that motions are not only noised with random noise $\sqrt{(1 - \bar{\alpha}_{N})} \bm{\epsilon}$ but also shifted by the shifting parameter $\sqrt{\bar{\alpha}_{N}}$. Therefore, the distance at noise step $N$ becomes the following: 
\begin{equation}
    D_N (\bm{m}^{(i)}_0, \bm{m}^{(k_i)}_0, d) = \sqrt{\bar{\alpha}_{N}} D_0 (\bm{m}^{(i)}_0, \bm{m}^{(k_i)}_0, d). \label{equation: N to 0}
\end{equation}
If this distance $D_N (\bm{m}^{(i)}_0, \bm{m}^{(k_i)}_0, d)$ is equal to or lower than the standard deviation of the distribution of noised motion $\sqrt{(1 - \bar{\alpha}_{N})}$, the retrieved motion $\bm{m}^{(k_i)}_0$ is within the distribution of noised motion $\mathcal{N}(\bm{m}_N^{(i)}; \sqrt{\bar{\alpha}_{N}} \bm{m}_{0}^{(i)}, (1 - \bar{\alpha}_{N}) \bf{I})$ at the noise step $n$. Therefore, $\bar{\alpha}_{N}$ should satisfy the following inequality:
\begin{equation}
    D_N (\bm{m}^{(i)}_0, \bm{m}^{(k_i)}_0, d) \leq \sqrt{(1 - \bar{\alpha}_{N})} \label{equation: inequality}
\end{equation}
Since we would like to minimize the noise scale $\sqrt{(1 - \bar{\alpha}_{N})}$ while satisfying Eq. (\ref{equation: inequality}) for all motions and dimensions, $\bar{\alpha}_{N}$ is derived from Eqs.~(\ref{equation: N to 0}) and (\ref{equation: inequality}) as follows:
\begin{equation}
    \bar{\alpha}_{N} = \frac{1}{1+ \max\limits_{\substack{i \in \{1,\cdots,J\} \\ d \in \{1,\cdots,d_m\}}} D_0(\bm{m}^{(i)}_0, \bm{m}^{(k_i)}_0, d)}
\end{equation}
Finally, the parameters ($\beta_{1},\cdots,\beta_{N}$) are optimized. As in~\cite{ho2020denoising}, $\beta_{t}$ is linearly increased from $\beta_{0}$ as follows:
\begin{equation}
    \beta_{n} = \beta_{0} + \gamma n
\end{equation}
$\beta_{0}$ is set to a small value for accurately refining the motion at early noise steps, and $\gamma$ controls the increment of the noise scale. Since $\bar{\alpha}_{N} = \prod_{n=1}^{N} (1 - \beta_{n})$, $\gamma$ is derived by solving the following equation: 
\begin{equation}
    \prod_{n=1}^{N} (1 - \beta_{0} - \gamma n) = \frac{1}{1+ \max\limits_{\substack{i \in \{1,\cdots,J\} \\ d \in \{1,\cdots,d_m\}}} D_0(\bm{m}^{(i)}_0, \bm{m}^{(k_i)}_0, d)} \label{equation: gamma}
\end{equation}
Since it is difficult to derive a closed-form solution for Eq.~(\ref{equation: gamma}), $\gamma$ is obtained by traditional numerical optimization in this paper.
While there are many numerical optimization methods, we use gradient descent because of its simplicity in implementation.

\subsection{STE retrieval}
\label{section:STE based retrieval}

\begin{figure}[h]
  \begin{center}
    \includegraphics[width=1.0\columnwidth]{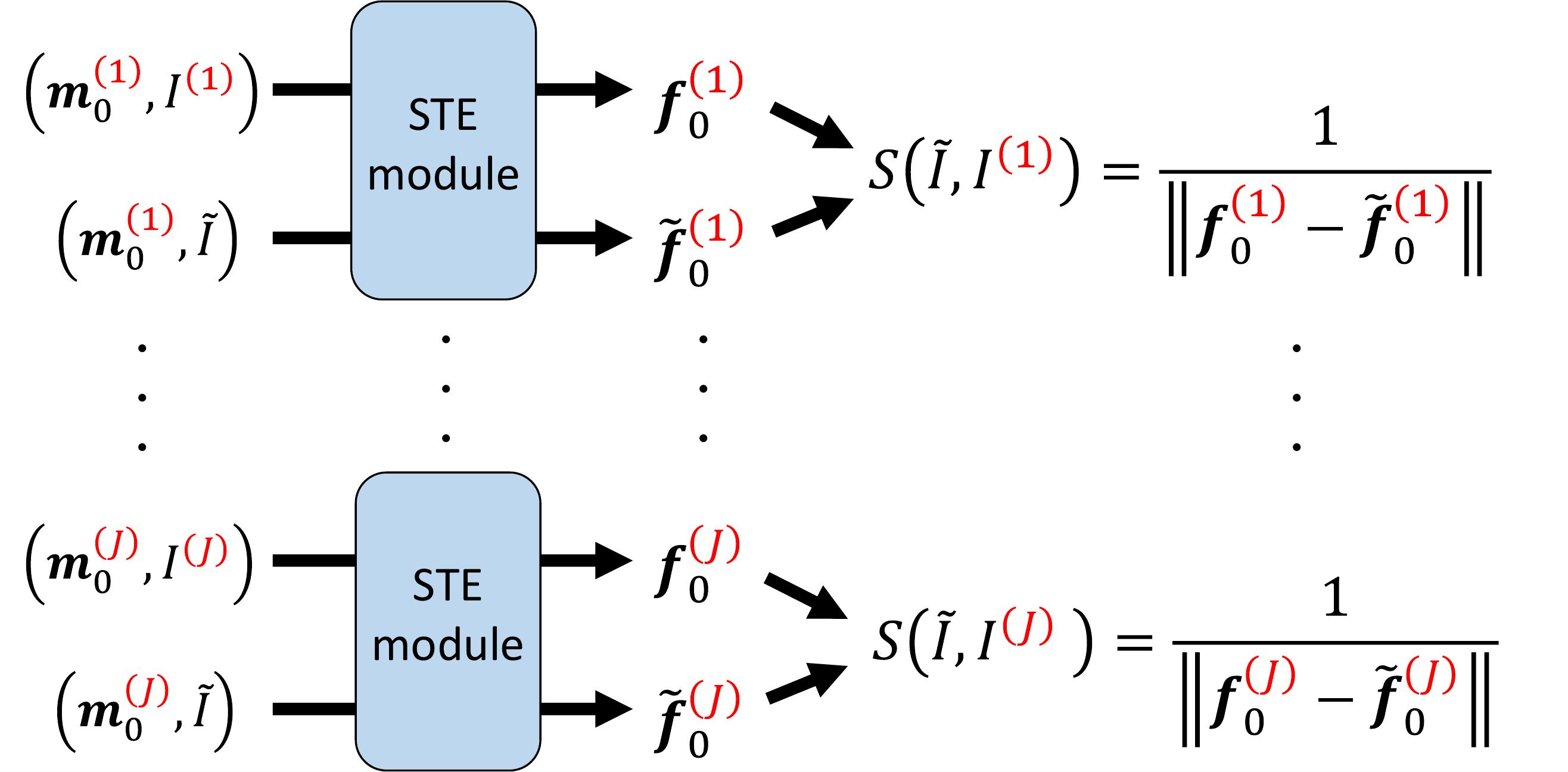}
    \caption{Similarity computation for STE retrieval. The details of the STE module are explained in Section~\ref{section:model architecture}.}
    \label{fig:Retrieval}
  \end{center}
\end{figure}

This section introduces our image-based motion retrieval method, as shown in the right side of Fig.~\ref{fig:procedure of R2-Diff}.
Motions are retrieved based on the similarity of image features in the inference stage of R2-Diff. Since our diffusion model is trained to denoise the motion based on the image features extracted by the STE module, these image features should contain important information for motion prediction, such as object position and geometry. This information is important for image-based motion retrieval, and thus STE features are also effective for retrieval. STE, however, requires the motion to extract the image feature, although the corresponding motion for the test image is unknown. To solve this problem, the motion $\bm{m}_0^{(i)}$ corresponding to each image $I^{(i)}$ in the training dataset is applied to the test image $\tilde{I}$, extracting different features for each $i$.
Let's consider computing the similarity of images $I^{(i)}$ and $\tilde{I}$. Figure~\ref{fig:Retrieval} shows the procedure for computing the similarity. First, the image feature for $I^{(i)}$ is obtained by inputting $I^{(i)}$ and $\bm{m}_0^{(i)}$ into the STE module. This image feature is denoted as $\bm{f}^{(i)}_0$ in Fig.~\ref{fig:Retrieval}. Then, the image feature for $\tilde{I}$, denoted as $\tilde{\bm{f}}^{(i)}_0$, is obtained by inputting $\tilde{I}$ and $\bm{m}_0^{(i)}$ into the STE module. This is because the most important feature for $I^{(i)}$ is along the trajectory of the motion $\bm{m}_0^{(i)}$. In other words, other regions do not contain important features. Therefore, the image features $\bm{f}^{(i)}_0$ and $\tilde{\bm{f}}^{(i)}_0$ along the motion are sufficient to compute the similarity of the images $I^{(i)}$ and $\tilde{I}$. After extracting the image features as mentioned above, the similarity $S(\tilde{I}, I^{(i)})$ is computed as follows:
\begin{equation}
    S(\tilde{I}, I^{(i)}) = \frac{1}{\| \bm{f}^{(i)}_0 - \tilde{\bm{f}}^{(i)}_0 \|}
\end{equation}
The motion is retrieved by selecting the index $i$ with the highest image similarity.

\section{Experiments}
\label{section: experiments}
\subsection{Dataset and Evaluation metric}

\begin{figure*}[t]
 \begin{center}
    \includegraphics[width=1.0\columnwidth]{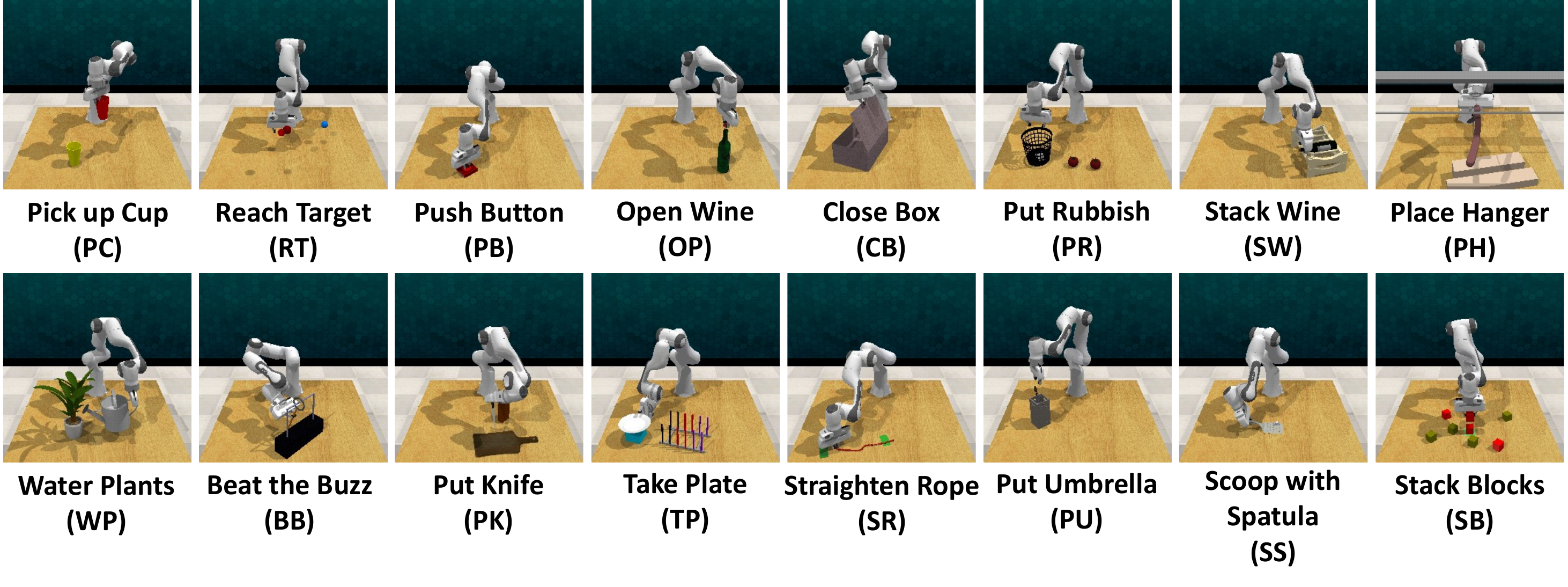}
    \caption{Task list of our experiment. The camera position and the robot's initial position are fixed. The other objects are randomly placed on the desk.}
    \label{fig:dataset}
 \end{center}
\end{figure*}

We evaluated our method on sixteen different robot manipulation tasks provided from RLBench~\cite{james2019rlbench}.
The list of the tasks is shown in Fig.~\ref{fig:dataset}.
We prepared 1,000 and 100 sequences for training and testing, respectively.
The motion length $T$ is aligned to 100 for all tasks.
The success rate of predicted motions on the test data is used to evaluate the performance. 
The predicted motion is considered a success if the reward given by RLBench is $1$ or more. This reward is given when the task is accomplished. Otherwise, the motion is regarded as a failure.

\subsection{Comparison with SOTA methods}
\label{section:Comparison Methods}
We evaluate the success rate of the following six methods to compare R2-Diff with state-of-the-art methods.

\noindent
{\bf (a) RT1 (from scratch):} RT1~\cite{rt12022arxiv} is one of the state-of-the-art classification-based motion prediction methods. The motion is discretized to $k$-bins, and RT1 predicts the index of the bin for each frame at once. While the original RT1 achieved state-of-the-art performance by optimizing the model with a huge dataset, RT1 is optimized from scratch with a single task for a fair comparison in our scenario (with a limited number of data).

\noindent
{\bf (b) VINN:} VINN~\cite{pari2021surprising} is one of the retrieval-based motion prediction methods. VINN retrieves motions from the dataset by utilizing self-contrastive learning models. As in~\cite{pari2021surprising}, the retrieval model is trained by Bootstrap Your Own Latent ~\cite{grill2020bootstrap} (BYOL) with our dataset. Moreover, we experimentally found that group normalization~\cite{wu2018group} is better than batch normalization~\cite{santurkar2018does} for our dataset (see Section~\ref{section:evaluation of retrieval}). Therefore, we evaluate VINN with group normalization in our experiments.

\noindent
{\bf (c) DMO-EBM:} DMO-EBM~\cite{oba2023data} is a refinement-based motion prediction model. As in our method, DMO-EBM predicts the motion appropriate to the image by refining the retrieved motion. The differences between R2-Diff and DMO-EBM are as follows: (1) While R2-Diff refines the motion based on the reverse diffusion process, DMO-EBM refines the motion using the simple regression model named Deep motion optimizer (DMO). (2) R2-Diff retrieves the motion based on the similarity of the STE feature, as explained in Section~\ref{section:STE based retrieval}. In contrast, DMO-EBM retrieves the motion based on the predicted probability density. This density is predicted by an energy-based model (EBM) employing the STE module for feature extraction. In~\cite{oba2023data}, this EBM is optimized by maximum likelihood estimation that is less stable than the score-based method (e.g., denoising score matching~\cite{song2019generative}, diffusion~\cite{ho2020denoising}).

\noindent
{\bf (d) Diffusion Policy (Diff-Policy):} Diffusion policy~\cite{chi2023diffusion} is a diffusion-based motion prediction method. Diffusion policy differs from R2-Diff in the prediction method, tuning, and model architecture. As in the traditional diffusion models, diffusion policy predicts the motion without utilizing retrieval by denoising the motion sampled from the standard normal distribution in diffusion policy, and thus the parameters $\beta_1,\cdots,\beta_{N}$ are tuned so that $\bar{\alpha}_{N}$ becomes $1$. Regarding image feature extraction, the diffusion policy employs the Resnet that extracts the image features independent of the motion, while R2-Diff employs the STE module that extracts the image features along the motion. Different from our problem setting, the original diffusion policy predicts the short length of motion and repeats updating the motion after executing the predicted motion. To fit our problem setting, we changed the last layer of the original diffusion policy model so that the model predicts the whole motion at once.  

\noindent
{\bf (e) R2-Diff:} Our proposed diffusion-based model. The total denoising step $N$ and step for adding noise to retrieved motion $n$ are set to 1000 and 500 in all experiments in Section~\ref{section: experiments} unless specified, respectively. 

\begin{table}[!t]
\caption{Comparison of the success rate of each manipulation task between our proposed method and recent state-of-the-art methods.
``Avg'' in this table denotes the average success rate across all tasks.
}
\begin{center}
\scalebox{0.95}{
\begin{tabular}{l|c c c c c c c c c}
\hline
               & Avg         & PC      & RT      & PB       & OW      & CB      & PR     & SW     & PH \\ \hline
RT1            & 20.5        &  0        & 83      & 0         & 21         & 0        & 0        & 0         & 0     \\ \hline
VINN           & 18.7       & 5         & 2        & 2         & 20         & 59      & 2        & 45       & 4     \\ \hline
DMO-EBM        & 56.6       & 89       & 32      & 85      & 61          & 97      & 85      & 74       & 40   \\ \hline
Diff-Policy    & 46.0       & 2          & 0        & 1         & 77         & 99      & 100    & 100    & 40    \\ \hline
R2-Diff (ours) & 62.9   & 95        & 91      & 99       & 72         & 96      & 98      & 96      & 43    \\ \hline \hline
                      &      -        & WP      & BB     & PK      & TP         & SR      & PU     & SS      & SB    \\ \hline
RT1               &      -        & 66        & 65      & 7         & 86         & 0        & 0        & 0         & 0     \\ \hline
VINN            &      -        & 71        & 25      & 25       & 38         & 0        & 1        & 0         & 0     \\ \hline
DMO-EBM   &      -        & 91        & 46      & 51       & 94         & 16      & 25      & 2        & 17   \\ \hline
Diff-Policy   &      -        & 92        & 95      & 3         & 98         & 0        & 28      & 2        & 0    \\ \hline
R2-Diff (ours) &   -       & 90        & 56      & 48       & 89         & 23      & 6        & 3         & 1      \\ \hline 
\end{tabular}
}
\end{center} \label{table:Comparison with SOTA}
\end{table}

Table~\ref{table:Comparison with SOTA} shows the success rates of each method. 
The average success rate for all the tasks labeled Avg in Table 1 shows that our proposed method has the highest success rate. The reasons for this are (1) the model structure and (2) the combination of retrieval and diffusion. First, the effectiveness of the model structure is validated by the fact that DMO-EBM and R2-Diff, which use STE modules, achieve a higher success rate than the other models. Then, the comparison of DMO-EBM and R2-Diff demonstrates that the combination of retrieval and diffusion is better than the combination of refinement with regression and retrieval with EBM. However, it is unclear how effective the combination of retrieval and diffusion is compared to traditional diffusion models. 
Therefore, in Section~\ref{section:evaluation with same arch}, we verify the effectiveness of retrieval by aligning their model architecture. 

\subsection{Effectiveness of retrieval and tuning for diffusion}
\label{section:evaluation with same arch}

\begin{table}[!t]
\caption{Comparison of the success rate of each manipulation task between our retrieval-based and traditional random-sample-based diffusions. The mode indicates the initial sample selection method. The initial sample is ``randomly selected from a standard normal distribution'' and ``retrieved with STE'' in ``rand'' and ``ret'', respectively. While ``Basic diffusion'' is optimized with traditional hyperparameters, our hyperparameter tuning is applied in ``R2-Diff''.}
\begin{center}
\scalebox{0.95}{
\begin{tabular}{l|l c c c c c c c c c}
\hline
                 & mode & Avg   & PC   & RT  & PB  & OW & CB   & PR  & SW  & PH \\ \hline
Basic diffusion  & rand & 54.4  & 96    & 1    & 44  & 70    & 100 & 96   & 98    & 40  \\ \hline
Basic diffusion  & ret  & 52.2  & 95    & 8    & 42  & 58    & 99   & 93   & 96    & 38   \\ \hline
R2-Diff (ours)   & rand & 21.6  & 54    & 0    & 1    & 86    & 3     & 15   & 100  & 18   \\ \hline
R2-Diff (ours)   & ret  & 62.9 & 95    & 91  & 99  & 72    & 96   & 98   & 96    & 43    \\ \hline \hline
                 & mode & -       & WP & BB  & PK  & TP  & SR  & PU    & SS    & SB    \\ \hline
Basic diffusion  & rand  &  -      & 84    & 84  & 52   & 98  & 0    & 1       & 6      & 1     \\ \hline
Basic diffusion  & ret      &  -      & 84   & 78  & 41   & 98   & 2    & 2      & 1       & 1     \\ \hline
R2-Diff (ours)   & rand   &  -      & 0     & 7    & 37    & 21  & 0     & 3     & 0      & 0   \\ \hline
R2-Diff (ours)   & ret      &   -     & 90   & 56   & 48   & 89  & 23   & 6      & 3      & 1      \\ \hline 
\end{tabular}
}
\end{center} \label{table:evaluation with same arch}
\end{table}
We evaluate traditional and proposed diffusion models by aligning the ``model architecture'' to fairly examine the effectiveness of our method. We employ the architecture explained in Section~\ref{section:model architecture}. 
Table~\ref{table:evaluation with same arch} shows their success rates. 
Basic diffusion in Table~\ref{table:evaluation with same arch} denotes the diffusion model with traditional hyper-parameters; $\beta_1, \cdots, \beta_{N}$ are linearly improved and tuned so that $\bar{\alpha}_{N}$ becomes 1 as in~\cite{ho2020denoising}. R2-Diff denotes $\beta_1, \cdots, \beta_{N}$ are tuned by the proposed method explained in Section~\ref{table:evaluation with same arch}. 
Rand and ret in Table~\ref{table:evaluation with same arch} denote the predictions from a random sample and a retrieved motion, respectively.

See the result of rand and ret of Diffusion in Table~\ref{table:evaluation with same arch}.
The average success rate of ret is lower than that of rand with the basic diffusion. On the other hand, ret with R2-Diff achieve the highest success rate among them, while rand with R2-Diff is the worst. These results reveal that our tuning is effective when the appropriate retrieved motion is fed into the model.

\subsection{Evaluation of rank for tuning }
\label{section:evaluation of rank}

\begin{table}[!t]
\caption{Comparison of the success rate of each manipulation task among R2-Diffs optimized with various hyperparameters computed based on the rank.}
\begin{center}
\scalebox{0.95}{
\begin{tabular}{l|l c c c c c c c c c}
\hline
                             & Avg   & PC   & RT  & PB  & OW & CB   & PR  & SW  & PH \\ \hline
R2-Diff (rank 1)  & 62.9 & 95    & 91  & 99   & 72    & 96   & 98   & 96    & 43   \\ \hline 
R2-Diff (rank 5)  & 62.1 & 95    & 91  & 93   & 74    & 96   & 100 & 100  & 49   \\ \hline
R2-Diff (rank 10) & 58.9 & 95    & 89  & 94   & 65    & 71   & 95   & 100  & 32     \\ \hline \hline
                             & -       & WP & BB  & PK  & TP  & SR  & PU    & SS    & SB    \\ \hline
R2-Diff (rank 1)  & -       & 90   & 56   & 48   & 89  & 23   & 6      & 3      & 1      \\ \hline
R2-Diff (rank 5)  & -       & 90   & 37   & 52   & 87  & 23   & 6      & 0      & 1     \\ \hline
R2-Diff (rank 10) & -       & 91   & 65   & 49   & 86  & 0     & 7      & 1      & 3     \\ \hline
\end{tabular}
}
\end{center} \label{table:success rate with different rank}
\end{table}

\begin{table}[!t]
\caption{The value of $\bar{\alpha_{N}}$ for various rank.}
\begin{center}
\scalebox{0.95}{
\begin{tabular}{l c c c c c c c c c}
\hline
                 & PC       & RT     & PB      & OW       & CB      & PR      & SW       & PH \\ \hline
R2-Diff (rank 1)  & 0.848    & 0.973  & 0.935   & 0.910    & 0.789   & 0.869   & 0.821    & 0.701   \\ \hline 
R2-Diff (rank 5)  & 0.779    & 0.904  & 0.907   & 0.841    & 0.798   & 0.707   & 0.775    & 0.566   \\ \hline
R2-Diff (rank 10) & 0.765    & 0.881  & 0.824   & 0.808    & 0.786   & 0.672   & 0.760    & 0.564     \\ \hline \hline
                 & WP      & BB        & PK        & TP      & SR        & PU           & SS           & SB    \\ \hline
R2-Diff (rank 1)  & 0.997   & 0.725   & 0.638   & 0.859  & 0.773   & 0.526      & 0.876      & 0.688      \\ \hline
R2-Diff (rank 5)  & 0.995   & 0.602   & 0.571   & 0.851  & 0.736   & 0.502      & 0.845      & 0.663     \\ \hline
R2-Diff (rank 10) & 0.993   & 0.583   & 0.532   & 0.855  & 0.738   & 0.473      & 0.688      & 0.620     \\ \hline
\end{tabular}
}
\end{center} \label{table:alpha with different rank}
\end{table}

In R2-Diff, the noise scale is tuned to fit the distance to the nearest neighbor motion based on the assumption that the nearest neighbor motion is retrieved in inference. However, this assumption is not always true because the nearest neighbor motion is not retrieved sometimes depending on the performance of the retrieval method. Therefore, we verify whether our tuning is appropriate by changing the target value of the noise scale from the distance of the 1st nearest neighbor to the $k$-th nearest neighbor. Based on the change of the rank $k$, $\bm{m}^{(k_i)}_0$ in Eq.~(\ref{equation: inequality}) is changed to the $k$-th nearest neighbor motion. 

Table~\ref{table:success rate with different rank} shows the success rates of R2-Diff with different $k$. We also show $\bar{\alpha}_{N}$ for different $k$ in Table~\ref{table:alpha with different rank}. As shown in Table~\ref{table:success rate with different rank}, $k=1$ is best in terms of the average success rate. Therefore, our tuning is best for our dataset. However, Table~\ref{table:success rate with different rank} validates that the optimal $k$ is different for each task. This is because the retrieval difficulty depends on the tasks, and thus $k$ should also be optimized for each task in future work. 

\subsection{Evaluation of retrieval methods}
\label{section:evaluation of retrieval}

 \begin{figure}[t]
  \begin{center}
    \includegraphics[width=1.0\columnwidth]{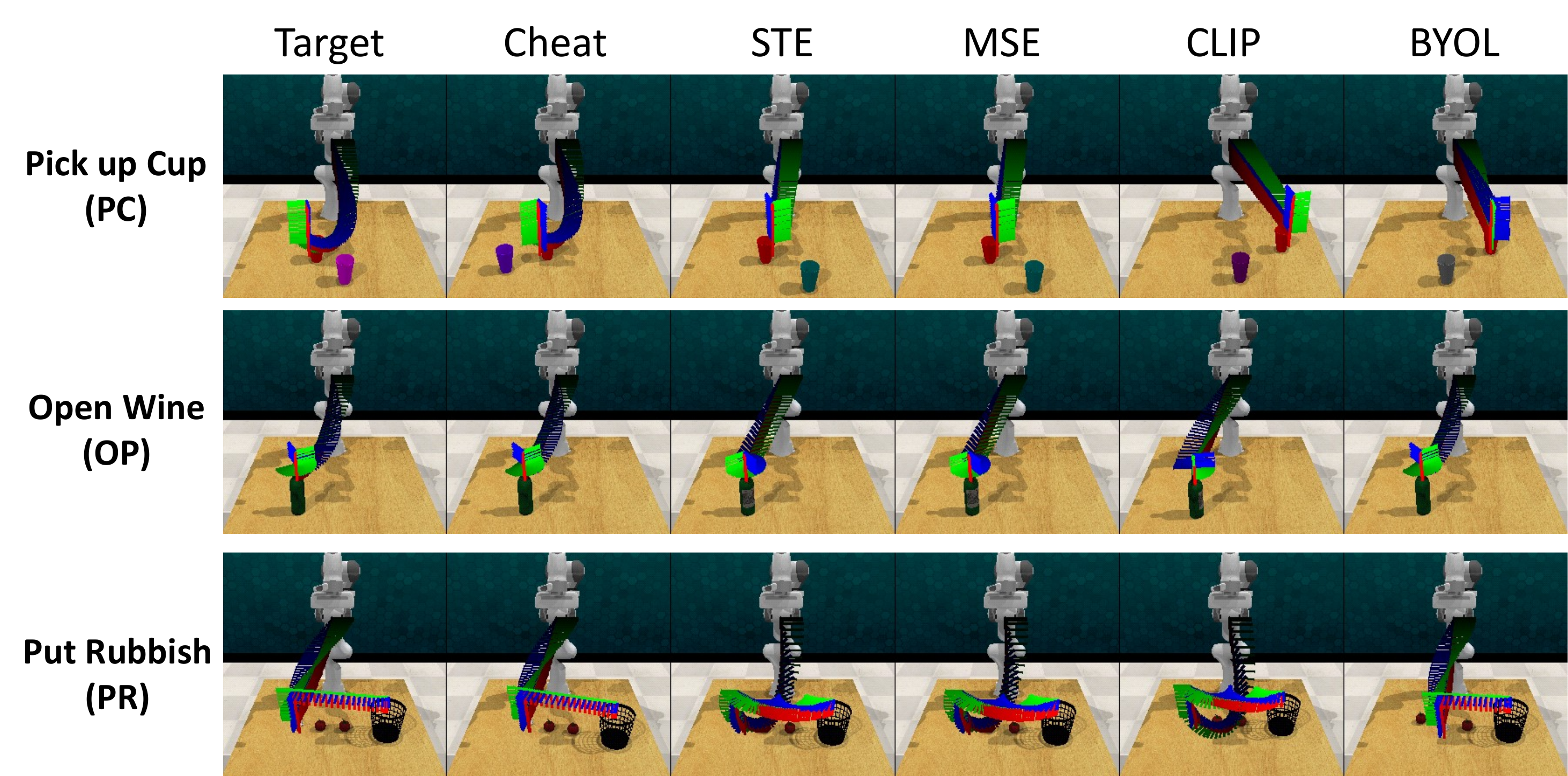}
    \caption{Examples of retrieved motion. The motions are retrieved to minimize the similarity of the image or motion of the target.}
    \label{fig:retrieval result}
  \end{center}
\end{figure}

\begin{table}[!t]
\caption{Comparison of the success rate of each manipulation task among various retrieval methods. The success rates are evaluated with retrieved motions without refinement.}
\begin{center}
\scalebox{0.95}{
\begin{tabular}{l|c c c c c c c c c}
\hline
               & Avg   & PC   & RT  & PB   & OW & CB   & PR  & SW  & PH \\ \hline
Cheat          & 60.6  & 85   & 96   & 96   & 57    & 86   & 99   & 85    & 42  \\ \hline
STE (rank 1)   & 58.5  & 95   & 75   & 100 & 76    & 93   & 98   & 87    & 34   \\ \hline
STE (rank 5)   & 56.8  & 95   & 74   & 100 & 83    & 94   & 99   & 80    & 36    \\ \hline
STE (rank 10)  & 57.1  & 96   & 78   & 100 & 74    & 73   & 99   & 89    & 31    \\ \hline
MSE (rank 1)   & 58.2  & 96   & 62   & 100 & 72    & 95   & 99   & 82    & 35    \\ \hline
MSE (rank 5)   & 56.6  & 94   & 71   & 100 & 78    & 97   & 100 & 77    & 32    \\ \hline
MSE (rank 10)  & 58.4  & 95   & 77  & 100  & 80    & 84   & 100 & 84    & 27    \\ \hline
BYOL-bn        & 14.1  & 12   & 3    & 7      & 34    & 27   & 6     & 26    & 0    \\ \hline
BYOL-gn        & 24.9  & 7     & 4    & 17    & 27    & 61   & 20   & 57    & 7    \\ \hline
CLIP           & 26.2  & 12   & 3    & 40    & 23    & 57   & 33   & 50    & 13    \\ \hline \hline
               & dims   & WP & BB  & PK  & TP   & SR  & PU    & SS    & SB    \\ \hline
Cheat          &  -     & 87   & 76  & 51   & 94   & 1     & 13     & 0      & 1  \\ \hline
STE (rank 1)   & 128*100& 93   & 50  & 38   & 89   & 1     & 5       & 1      & 1   \\ \hline
STE (rank 5)   & 128*100& 87   & 34  & 40   & 81   & 1     & 4       & 1      & 0     \\ \hline
STE (rank 10)  & 128*100& 92   & 62  & 37   & 79   & 2     & 1       & 0      & 0    \\ \hline
MSE (rank 1)   & 128*256*256    & 89   & 55  & 43   & 92   & 2     & 9       & 1      & 0    \\ \hline
MSE (rank 5)   & 128*256*256& 86   & 36  & 38   & 89   & 2     & 5       & 1      & 1    \\ \hline
MSE (rank 10)  & 128*256*256& 81   & 61  & 42   & 93   & 1     & 6       & 0      & 4    \\ \hline
BYOL-bn        & 256    & 57   & 15  & 5     & 31    & 0     & 2      & 0      & 0    \\ \hline
BYOL-gn        & 256    & 76   & 37  & 29  & 53    & 0     & 2      & 1       & 1    \\ \hline
CLIP           & 256    & 77   & 39  & 21  & 44    & 0     & 5      & 2       & 0    \\ \hline \hline
\end{tabular}
}
\end{center} \label{table:evaluation of retrieval}
\end{table}

This section examines the performance of retrieval methods. Table~\ref{table:evaluation of retrieval} shows the success rates of each task when the retrieved motions from various retrieval methods are executed without refinement. The retrieved motions are not refined to focus on evaluating the retrieval. Cheat in Table~\ref{table:evaluation of retrieval} is a baseline that the motion is retrieved based on the motion similarity, expressed in Eq.~(\ref{equation: motion retrieval}), with one of the ground-truth motions. Although the ground-truth motions are usually not available in inference, we prepared them in advance just for evaluation. The other methods retrieve the motion based on the similarity in image features. STE uses the STE features shown as in Fig.~\ref{fig:model}. MSE uses the whole image feature denoted as $F$ in Fig~\ref{fig:model}. The rank $k$ (e.g., R2-Diff rank 1) in Table~\ref{table:evaluation of retrieval} indicates the model is trained with the noise scale parameters $\beta$ are tuned based on the $k$-th nearest neighbor motion, and the model is trained with these parameters, as in Section~\ref{section:evaluation of rank}. BYOL-bn and BYOL-gn are the BYOL~\cite{grill2020bootstrap}, one of the self-contrastive learning methods, with batch and group normalization, respectively. Following the implementation in~\cite{pari2021surprising}, we use Resnet to compute the image features and optimize the model with our dataset. CLIP~\cite{radford2021learning} is one of the foundation models. Since the CLIP model successfully descript the contents of images from extracted image features, these image features possibly have information that is important for motion retrieval. To retrieve the motion by CLIP, We use the image feature obtained from Resnet pre-trained with the CLIP dataset. In addition to the success rates, the dimensionality of image features in these retrieval methods is also shown in Table~\ref{table:evaluation of retrieval}. The dimensionality affects the computational cost for retrieval. The experiment results of each method are explained in the following.

\noindent
{\bf BYOL~\cite{grill2020bootstrap}: }
The results of BYOL indicate that self-contrastive learning is not suitable for this dataset. Self-contrastive learning optimizes the model so that the image features of positive pairs become close to each other while those of negative pairs become far away. The positive image pairs are created by applying random augmentation (e.g., blur, noise) to the same index image. In contrast, the negative pairs are created by the different index images based on the assumption that the similarity of these images is low. 
This contrastive learning is effective for general image classification or retrieval. However, this method is not effective in the scenario of motion prediction for the robot, especially when the background image is almost the same. This is because the variety of images is less than the variety of motion in such a scenario. In this case, for example, contextually similar image pairs (e.g., the position of target objects is almost the same) exist in the dataset. While these image pairs should be treated as positive pair, BYOL treats these pairs as negative pair because the index of the image is different. Therefore, BYOL fails to learn the appropriate distance and retrieve the correct motion.

\noindent
{\bf CLIP~\cite{radford2021learning}: }
The CLIP feature is also not effective for our dataset. CLIP is trained so that the features from the image and sentence become close if the sentence appropriately describes the image. Therefore, the CLIP features are often used to retrieve images from a sentence. However, since the sentence for training CLIP ``roughly'' describes the position (e.g., left), it is difficult to retrieve the image by ``precise'' position information. For this reason, CLIP failed to retrieve the motion accurately. 

\noindent
{\bf STE and MSE}
Both STE and MSE achieve high success rates close to Cheat, regardless of the rank $k$. Interestingly, the per-task success rates for STE and MSE sometimes exceed those of Cheat. This is because there are multiple correct motions for an image. Since the motion is retrieved by the similarity of the single ground-truth motion in Cheat, Cheat cannot retrieve a correct motion that is completely different from the prepared ground-truth motion. On the other hand, STE and MSE successfully retrieved correct motions that differed from the prepared ground-truth motions, as shown in Fig.~\ref{fig:retrieval result}, and thus sometimes exceeded the success rate of Cheat. 
STE and MSE have similar success rates, although there is a large difference in dimensionality and thus in computational and memory costs, as shown in Table~\ref{table:evaluation of retrieval}. These results reveal that STE performs better relative to the computational cost.

\subsection{Evaluation of refinement}

\begin{table}[!t]
\caption{Comparison of the success rates of each manipulation task among various refinement methods. 
}
\begin{center}
\scalebox{0.95}{
\begin{tabular}{l|l c c c c c c c c c}
\hline
                 & Avg   & PC   & RT  & PB  & OW    & CB   & PR  & SW  & PH \\ \hline
w/o refinement   & 60.6  & 85   & 96  & 96  & 57    & 86   & 99   & 85    & 42  \\ \hline 
Basic diffusion  & 55.6  & 100  & 1   & 44  & 78    & 100  & 98   & 96    & 44  \\ \hline 
DMO              & 58.5  & 99   & 0   & 89  & 57    & 83   & 100  & 86    & 39     \\ \hline 
R2-Diff (rank1)  & 70.4  & 100  & 96  & 98   & 85   & 99   & 100  & 98   & 55   \\ \hline 
R2-Diff (rank5)  & 69.6  & 98   & 95  & 90   & 82   & 100  & 98   & 99   & 62    \\ \hline
R2-Diff (rank10) & 66.8  & 99   & 90  & 95   & 81   & 99    & 99   & 100 & 50     \\ \hline \hline
                 & -     & WP & BB  & PK  & TP  & SR  & PU    & SS    & SB    \\ \hline
w/o refinement   & -     & 87   & 76   & 51  & 94   & 1    & 13     & 0      & 1      \\ \hline 
Basic diffusion  & -     & 83   & 83   & 59  & 98   & 3    & 2       & 1     & 0      \\ \hline 
DMO              & -     & 89   & 71   & 84  & 96   & 8    & 27     & 2      & 6     \\ \hline 
R2-Diff (rank1)  & -     & 87   & 78   & 79  & 100  & 19   & 32    & 1      & 1      \\ \hline
R2-Diff (rank5)  & -     & 91   & 76   & 74  & 100  & 33   & 13    & 2      & 0     \\ \hline
R2-Diff (rank10)& -      & 83   & 82   & 77  & 98   & 0    & 12    & 0     & 3     \\ \hline
\end{tabular}
}
\end{center} \label{table:comparison of refinement}
\end{table}

\begin{figure}[h]
  \begin{center}
    \includegraphics[width=1.0\columnwidth]{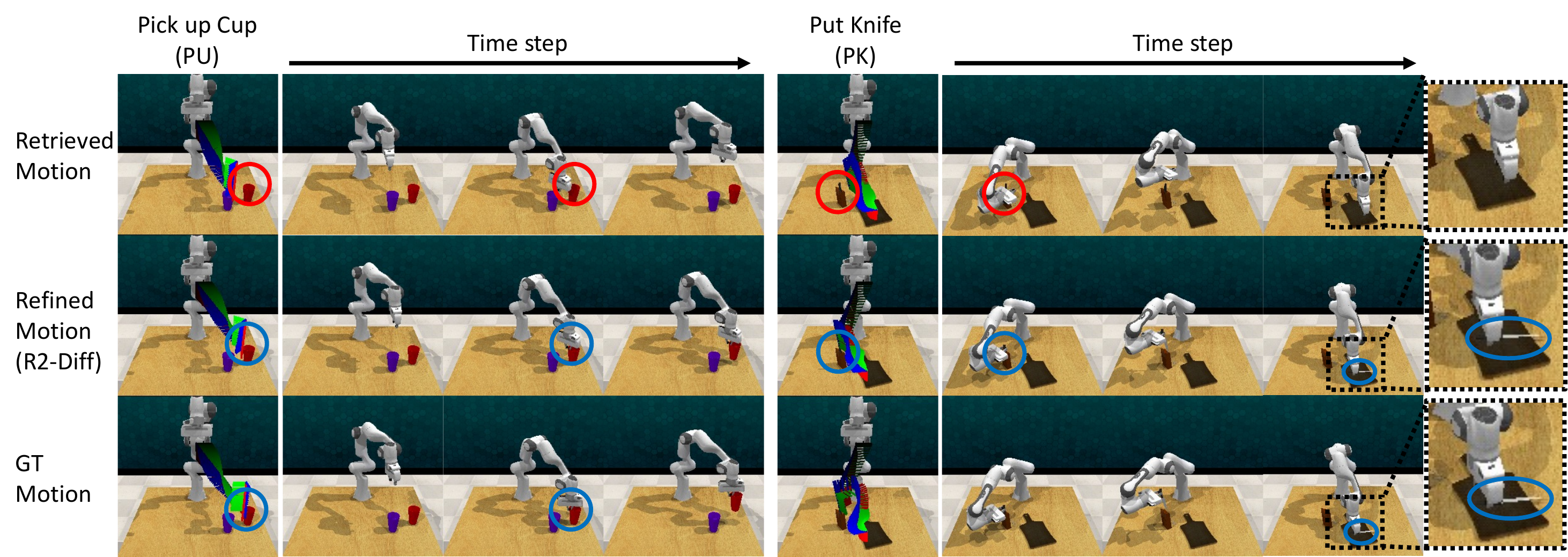}
    \caption{Examples of R2-Diff motion refinement. The circles in this figure indicate the important regions to check.}
    \label{fig:refinement example}
  \end{center}
\end{figure}

To evaluate the refinement performance, we compare R2-Diff with normal diffusion and DMO (refinement model in~\cite{oba2023data}). To properly evaluate only the refinement, we retrieved the motion closest to the correct motion prepared in advance, as in Cheat in Section~\ref{section:evaluation of retrieval}, and then refined the retrieved motion by each refinement method. In ``w/o refinement'' in Table~\ref{table:comparison of refinement}, refinement is not executed so that this method is identical to Cheat in Section~\ref{section:evaluation of retrieval}.

As shown in Table~\ref{table:comparison of refinement}, R2-Diff has a higher success rate than the other methods and increases the success rate from w/o refinement in most tasks. R2-Diff uses the nearest neighbor search between training data to estimate the error between the correct and retrieved motion, and adjusts parameters based on the estimated error to train the model. Therefore, the high performance of R2-Diff validates that learning based on the estimated errors of retrieved motion is important for effectively training the model.

\subsection{Evaluation of max and retrieval steps}
\label{section:evaluation noise step}

In this section, we examine the performance of R2-Diff by varying the number of noise steps $n$ applied to the retrieved motion and the overall noise step $N$. Table~\ref{table:change n} shows the success rates of R2-Diff with various $n$ when $N=1000$.

As shown in Table~\ref{table:change n}, the highest success rate is obtained with $n = 500$, although the change in success rate is small. This is because the noise scale is tuned to match the ``maximum'' of the error between the nearest neighbor motions when $n=N$. Since the average error is smaller than the maximum, $n=500$ is superior on average. 

Table~\ref{table:change N} shows the performance variation with varying $N$ when $n$ is fixed at $2/N$. As with conventional diffusion, the performance degraded as $N$ is reduced.

\begin{table}[!t]
\caption{Comparison of the success rate of each manipulation task among R2-diff with various $n$.}
\begin{center}
\scalebox{0.95}{
\begin{tabular}{l|l c c c c c c c c c}
\hline
                   & Avg   & PC   & RT  & PB  & OW & CB   & PR  & SW & PH \\ \hline
R2-Diff ($n=0$)    & 58.5  & 95   & 75  & 100 & 76   & 93    & 98   & 87   & 34   \\ \hline 
R2-Diff ($n=250$)  & 62.8  & 96   & 92  & 99   & 73   & 96    & 98   & 100 & 43    \\ \hline
R2-Diff ($n=500$)  & 62.9  & 95   & 91  & 99   & 72   & 96    & 98   & 96   & 43    \\ \hline
R2-Diff ($n=750$)  & 62.1  & 94   & 92  & 97   & 69   & 96    & 97   & 98   & 40    \\ \hline
R2-Diff ($n=1000$) & 62.0  & 97   & 91  & 97   & 77   & 96   & 95   & 96   & 36     \\ \hline \hline
                   &  -    & WP   & BB  & PK  & TP  & SR  & PU    & SS    & SB    \\ \hline
R2-Diff ($n=0$)    & -     & 93   & 50   & 38   & 89   & 1    & 5       & 1     & 1       \\ \hline 
R2-Diff ($n=250$)  & -     & 93   & 50   & 52   & 87   & 17  & 5       & 2     & 1       \\ \hline
R2-Diff ($n=500$)  & -     & 90   & 56   & 48   & 89   & 23  & 6       & 3     & 1    \\ \hline
R2-Diff ($n=750$)  & -     & 92   & 54   & 51   & 90   & 12  & 10     & 1     & 1    \\ \hline
R2-Diff ($n=1000$) & -     & 94   & 50   & 46   & 90   & 18  & 8       & 0    & 1     \\ \hline
\end{tabular}
}
\end{center} \label{table:change n}
\end{table}

\begin{table}[!t]
\caption{Comparison of the success rate of each manipulation task among R2-diff with various $N$.}
\begin{center}
\scalebox{0.95}{
\begin{tabular}{l|l c c c c c c c c c}
\hline
                                   & Avg   & PC   & RT  & PB  & OW & CB   & PR  & SW & PH \\ \hline
R2-Diff ($N=100$)  & 56.4  & 96   & 86  & 99   & 72   & 87    & 96   & 100 & 37   \\ \hline 
R2-Diff ($N=500$)  & 59.2  & 94   & 99  & 87   & 78   & 97    & 85   & 100 & 27    \\ \hline
R2-Diff ($N=1000$)& 62.9  & 95   & 91  & 99   & 72   & 96    & 98   & 96   & 43    \\ \hline \hline
                                   &  -        & WP & BB  & PK  & TP  & SR   & PU    & SS    & SB    \\ \hline
R2-Diff ($N=100$)  & -        & 52   & 26   & 52   & 92   & 2    & 2       & 3     & 2    \\ \hline
R2-Diff ($N=500$)  & -        & 92   & 27   & 44   & 87   & 15  & 10     & 1     & 4    \\ \hline
R2-Diff ($N=1000$)& -        & 90  & 56   & 48    & 89   & 23  & 6       & 3     & 1     \\ \hline
\end{tabular}
}
\end{center} \label{table:change N}
\end{table}

\section{Limitations and future work}
One of the problems of R2-Diff is that refinement sometimes decreases the success rate. Table~\ref{table:before and after} shows the success rates of R2-Diff before and after the refinement. Table~\ref{table:before and after} shows that the refinement decreases the success rates in OW and WP. Interestingly, in these tasks, the success rate of motions retrieved by STE exceeds that of Cheat. Furthermore, in CB and SW, STE retrieval outperforms Cheat before refinement, but Cheat has higher success rates after refinement. In addition, the average success rates of STE and Cheat are almost the same before refinement, but the success rates of Cheat exceed those of STE by almost 8 percent after refinement. This is because the distribution of the distance between correct and retrieved motions, which was estimated by the nearest neighbor search in the training phase as in Section~\ref{section:scaling of noise}, is different from its real distribution. The proposed method sample noised motions from the Gaussian distribution whose standard deviation is the maximum distance between the correct motion and its nearest neighbor in the training data. These noised motions are used as a substitute for retrieved motion, and the model learns the refinement by predicting the noise in these noised motions. However, there are cases where a retrieved motion is out of the Gaussian distribution we used. For example, retrieved motion that does not require refinement, i.e., almost zero noise, is out of distribution. This is because, in high-dimensional spaces, noise whose norm is nearly zero has a low probability of being sampled from the Gaussian distribution. Then, the model fails to refine such motions. This issue should be addressed in future work to improve the success rates further.

\begin{table}[!t]
\caption{Comparison of the success rates of each manipulation task among various R2-diffs.}
\begin{center}
\scalebox{0.95}{
\begin{tabular}{l|l c c c c c c c c c}
\hline
Retrieval & refinement & Avg   & PC   & RT  & PB  & OW  & CB   & PR   & SW   & PH \\ \hline
Cheat     & before     & 60.6  & 85   & 96  & 96  & 57  & 86   & 99   & 85   & 42 \\ \hline
Cheat     & after      & 70.4  & 100  & 96  & 98  & 85  & 99   & 100  & 98   & 55 \\ \hline
STE       & before     & 58.5  & 95   & 75  & 100 & 76  & 93   & 98   & 87   & 34 \\ \hline
STE       & after      & 62.9  & 95   & 91  & 99  & 72  & 96   & 98   & 96   & 43 \\ \hline \hline
Retrieval & refinement & -     & WP   & BB  & PK  & TP  & SR  & PU    & SS   & SB \\ \hline
Cheat     & before     & -     & 87   & 76  & 51  & 94  & 1   & 13    & 0    & 1  \\ \hline
Cheat     & after      & -     & 87   & 78  & 79  & 100 & 19  & 32    & 1    & 1  \\ \hline
STE       & before     & -     & 93   & 50  & 38  & 89  & 1   & 5     & 1    & 1  \\ \hline
STE       & after      & -     & 90   & 56  & 48  & 89  & 23  & 6     & 3    & 1  \\ \hline 
\end{tabular}
}
\end{center} \label{table:before and after}
\end{table}

\section{Conclusion}
\label{section:conclusion}

We proposed R2-Diff, an image-based motion prediction method that uses diffusion models to refine a retrieved motion. Traditional diffusion fails to predict contextually appropriate motions because the initial motion is sampled from a standard normal distribution independent of the image context. In contrast, R2-Diff solves the above problem by employing a motion that is retrieved based on image similarity as the initial motion. Furthermore, we proposed how to tune the hyperparameters of diffusion to optimize the diffusion model for motion refinement. We demonstrated that our tuning effectively optimizes the model for motion refinement compared to traditional tuning by evaluating the success rates in various robot manipulation tasks. We also proposed the image similarity-based motion retrieval method, which efficiently computes the similarity based on the image features along the motion trajectory. Experimental results show that R2-Diff outperforms the recent state-of-the-art models thanks to these novelties.

However, while this paper reveals that the tuning of diffusion models for refinement is promising, our tuning possibly does not reflect actual retrieved motion distributions. 
Therefore, we will work on adjusting the tuning to more accurately reproduce the distribution of retrieved motions for future work.





\clearpage
\bibliographystyle{elsarticle-num}
\bibliography{submit}



\end{document}